\documentclass[1p]{elsarticle}

\usepackage{lineno,hyperref,amsmath}
\usepackage{url}
\usepackage{graphicx}
\usepackage{epstopdf,amsmath,color,appendix}
\modulolinenumbers[5]

\journal{Journal of \LaTeX\ Templates}

\begin{document}

\begin{frontmatter}

\title{Classification of Occluded Objects using Fast Recurrent Processing}

\author[mymainaddress]{Ozgur Yilmaz\corref{mycorrespondingauthor}}
\cortext[mycorrespondingauthor]{Corresponding Author}
\ead{ozyilmaz@turgutozal.edu.tr, Website: ozguryilmazresearch.org}

\address[mymainaddress]{Turgut Ozal University, Department of Computer Engineering, Ankara Turkey}

\begin{abstract}
Recurrent neural networks are powerful tools for handling incomplete data problems in computer vision, thanks to their significant generative capabilities. However, the computational demand for these algorithms is too high to work in real time, without specialized hardware or software solutions. In this paper, we propose a framework for augmenting recurrent processing capabilities into a feedforward network without sacrificing much from computational efficiency. We assume a mixture model and generate samples of the last hidden layer according to the class decisions of the output layer, modify the hidden layer activity using the samples, and propagate to lower layers. For visual occlusion problem, the iterative procedure emulates feedforward-feedback loop, filling-in the missing hidden layer activity with meaningful representations. The proposed algorithm is tested on a widely used dataset, and shown to achieve 2$\times$ improvement in classification accuracy for occluded objects. When compared to Restricted Boltzmann Machines, our algorithm shows superior performance for occluded object classification. 
\end{abstract}


\end{frontmatter}

\section{Introduction}
\label{intro}
\begin{figure*}
\begin{center}
\fbox{\includegraphics[width=0.9\textwidth]{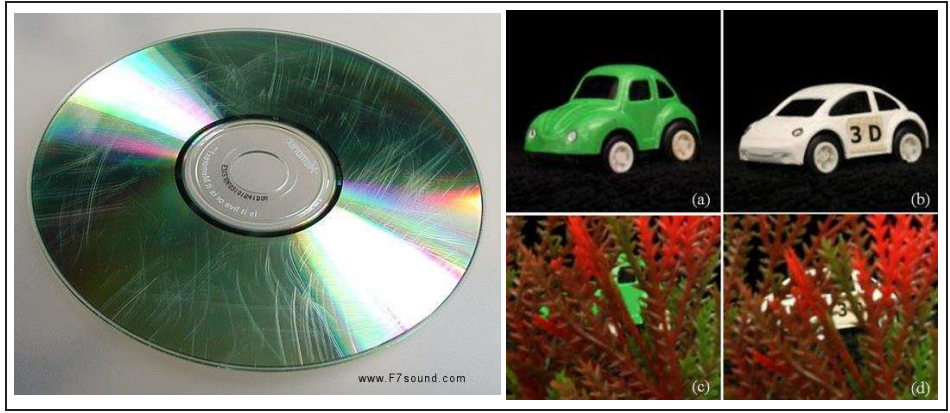}} 
\end{center}
   \caption{Incomplete data problem arises when you try to read from a scratched disk or recognize objects under occlusion. Adapted from www.f7sound.com and imagebank.osa.org}
\label{fig:fig00}
\end{figure*}

In many applications from a wide variety of fields, the data to be processed can partially be affected by severe noise in several phases, e.g., occlusions during a visual recording, a scratch on a compact disc or packet losses during transmission in a communication channel (Figure \ref{fig:fig00}). Missing data problems have been extensively studied for various different purposes (see \cite{garcia2010pattern} for a recent review). Such problems often severely degrade the performance of the target application; for instance pedestrian detection under occlusion \cite{dollar2012pedestrian} or face recogntion \cite{BurgosArtizzuICCV13rcpr}. 
 
Classification of objects under occlusions is an important problem in computer vision that has been previously tackled from both the cognitive \cite{johnson2005recognition,o2013recurrent,bornschein2013v1} and computational perspectives \cite{salakhutdinov2012efficient,chechik2008max,marlin2008missing,marlin2011recommender}. The proposed solutions are in the domain of inference with incomplete data \cite{rubin2009multiple}, generative models for classification problems \cite{smolensky1986information} and recurrent neural networks. 

In general incomplete data classification setting, the missing data attribute is known, but it is not the case in visual occlusions: detecting and localizing occluded region is not an easy task. In this paper, a novel algorithm based on mixture models \cite{dempster1977maximum}, multiple  \cite{rubin2009multiple} and K-Nearest Neighbor imputation \cite{batista2003experimental} is proposed. In our algorithm, neural network \cite{coates2011analysis} hidden layer activities are imputed in a recurrent architecture. The algorithm is tested on occluded object classification, where the occlusion pattern is completely unknown and its effect on the hidden layer activity is distributed. Our approach does not attempt to localize the occlusion but mitigate its effect on classification. 

The proposed multiple imputation approach can be considered as a pseudo-recurrent processing in the network as if it is governed by the dynamical equations of the corresponding hidden layer activities. This framework provides a shortcut into the feedback computation, which is suitable for real-time operation. The experiments show that the proposed algorithm improves classification performance significantly. When compared to Restricted Boltzmann Machines, our algorithm shows much superior performance and we conclude that energy based recurrent neural networks seem to be beneficial only when the occlusion is successfully localized. In the following, after discussing the related work in the literature, we describe the details of the proposed algorithm and present results on modified CIFAR 10 dataset \cite{krizhevsky2009learning}.

\subsection{Recurrent Neural Networks}
\label{sec:2}
Recurrent Neural Networks (RNNs) are connectionist computational models that utilize distributed representation and nonlinear dynamics of its units. Information in RNNs is propagated and processed in time through the states of its hidden units, which make them appropriate tools for sequential information processing. There are two broad types of RNNs: stochastic energy based RNNs with symmetric connections, and deterministic ones with directed connections. 
 
RNNs are known to be Turing complete computational models \cite{siegelmann1995computational} and universal approximators of dynamical systems \cite{funahashi1993approximation}. They are especially powerful tools in dealing with the long-range statistical relationships in a wide variety of applications ranging from natural language processing, to financial data analysis. 

Despite their immense potential as universal computers, difficulties in training RNNs arise due to the inherent problems of learning long-term dependencies \cite{hochreiter1991untersuchungen,bengio1994learning,hochreiter1997long} and convergence issues \cite{doya1992bifurcations}. However, recent advances suggest promising approaches in overcoming these issues, such as using better nonlinear optimizers \cite{martens2011learning,xu2013robust}, adopting hybrid strategies \cite{nasr2012training} or utilizing a reservoir of coupled oscillators \cite{maass2002real,jaeger2001echo}. Nevertheless, RNNs remain to be computationally expensive in both the training as well as the test phases. 

RNNs are shown to be very successful generative models for data completion \cite{bengio1996recurrent} and specifically in vision \cite{salakhutdinov2012efficient}. Although feedforward neural networks \cite{sharpe1995dealing,julie2012attribute,rey2012fuzzy}, auto associative neural networks \cite{marwala2006fault} and self organizing maps \cite{fessant2002self} are also used for data completion tasks, recurrent neural networks are a more natural choice due to their great generative capabilities. Restricted Boltzmann Machines (RBM) are shown to very successfully fill in the missing pixels due to visual occlusions \cite{salakhutdinov2012efficient}, but their classification performance were not measured for standard datasets. 

The idea in this paper is to imitate recurrent processing in a feedforward network and exploit its power while avoiding the expensive energy minimization in training, or computationally heavy sampling in test. More importantly, in our study we assume that the location of occlusion is unknown as opposed to \cite{salakhutdinov2012efficient,bengio1996recurrent}, which is the case in most real life applications (i.e. pedestrian detection \cite{dollar2012pedestrian}).

\subsection{Classification with Incomplete Data}
\label{sec:3}
Classification and clustering with missing data is a well-studied problem in the machine learning literature \cite{chechik2008max,marlin2008missing,wang2010classification,schafer2002missing} (see \cite{garcia2010pattern} for a review). The corresponding studies such as \cite{dempster1977maximum,rubin2009multiple,little2002statistical,smolensky1986information,batista2003analysis} are related to inference with incomplete data \cite{rubin2009multiple} and generative models \cite{smolensky1986information}, where Bayesian frameworks \cite{little2002statistical} are used for inference under missing data conditions. Alternatively, pseudo-likelihood \cite{besag1975statistical} and dependency network \cite{heckerman2001dependency} approaches solve data completion problem by learning conditional distributions. On the other hand, imputation is commonly used as a pre-processing tool \cite{little2002statistical}. The Mixture of Factor Analyzers \cite{ghahramani1994supervised} approach replaces the missing attributes with samples drawn from a parametric density, which models the distribution of the underlying true data. K-Nearest Neighbor imputation \cite{batista2003experimental} is shown to be a very effective method \cite{troyanskaya2001missing} despite its simplicity. 

Sampling from a mixture of factor analyzers \cite{ghahramani1994supervised} or the whole dataset \cite{batista2003experimental} and filling-in the missing data attributes is effectively very similar to the feedback information insertion in a neural network from a higher layer of neurons onto a lower layer of neurons. In this paper, imputation is used as a part of the pseudo-recurrent processing. Instead of imputing the missing data, the neural network hidden layer representation of the incomplete data is imputed in an iterative fashion, therefore we are proposing a novel framework. In our approach, a feedforward neural network makes a class decision at its output layer and based on this decision selects an appropriate density to estimate selected models' hidden layer activities. After this sampling stage, the algorithm inserts (weighted averaging) the estimated sample as if it is a feedback from a higher layer. This procedure is repeated multiple times to emulate the feedforward-feedback iterations in an RNN. Other related concepts such as multi-hypotheses feedback and winner-takes-all are also implemented to examine their role in this pseudo-recurrent processing. We suggest this approach as a real-time operable alternative to recurrent processing in neural networks. 

\subsection{Visual Occlusions}
\label{sec:1}
Data completion is essential in image processing for handling corrupted images. Generally, a corrupted image is restored by explicitly learning the image statistics \cite{weiss2007makes,lyu2007statistical} or by using neural networks \cite{gallinari1987memoires,jain2008natural,ranzato2013modeling}. These denoising studies treat incompleteness of images as a case of noise and filter it out using statistical approaches. Image inpainting \cite{bertalmio2000image} specifically deals with correcting structured noise in the image such as occlusions. A mask is provided by the user in inpainting application (as in the Restricted Boltzmann Machine study \cite{salakhutdinov2012efficient}) which pinpoints the occluded regions in the image, however a mask is almost never available in the general computer vision setting. 

On the other hand, there exist several studies that aim both localization and correction of occlusions. \cite{toh1990occlusion,yung1998detection,pang2004novel,zhang2008multilevel,yu2011robust,meger2011explicit}. In these studies, occlusion detection is performed using domain specific knowledge (visual cues) or external information (object geometry). However, these sources are not always available in general data imputation setting either. Other studies propose solutions via extracting occlusion maps using statistical measures. In \cite{wang2009hog}, HOG based classification errors; and in \cite{mei2011minimum}, template based reconstruction errors are used to generate such an occlusion map. In \cite{fransens2006mean}, a Hidden Markov Model (HMM) framework is utilized to estimate a visibility map, that localizes occlusions. 

To relieve the degrading effects of occlusion, part based models are also used: components are learned by imposing occlusion constraints in \cite{spratling2006learning}, descriptors are extracted from various parts of the occluded object in \cite{mohan2001example} and similarly; part-based descriptors are weighted with the occlusion measure in \cite{enzweiler2010multi}. Feature matching statistics are learned in \cite{ying2002partially}, visibility of face parts are learned in a cascade in \cite{BurgosArtizzuICCV13rcpr} and sub-matrix matching is utilized in \cite{saber2005partial} all of which can also be considered as part-based detection approaches. Learning occlusion patterns has also been investigated. In \cite{pepikj2013occlusion,gao2011segmentation,kwak2011learning} occlusion patterns are learned and localized in an object detection framework. \cite{hsiao2014occlusion} proposes a recurrent localization scheme, in which object detector hypothesizes about the occlusion using 3D geometry and top-down feedback corrects the hypothesis.

In general, occlusion localization is a computationally costly operation, and our approach aims at correction without localization. This is achieved by an iterative process that imitates a recurrent network and alleviates the degrading effect of missing data on classification. More importantly our classification approach is not part-based but holistic, in which localization of occlusion becomes more cumbersome. However, it should be noted that occlusion detection algorithms are very effective pre-processing tools for improving classification performance and our contribution is orthogonal to the literature on occlusion detection/localization.

\subsection{Contribution}
\label{sec:4}
We designed a real-time operable neural network based algorithm that has recurrence capabilities, suited for solving occlusion problem in object classification, without resorting to user defined occlusion masks or computationally expensive occlusion detection mechanisms. The proposed algorithm is a standard convolutional network augmented by feedback mechanisms, which is very fast and capable of classifying occluded objects with high accuracy. To our knowledge, this is the first neural network study that systematically examines the effect of occlusion on classification performance in a standard image dataset.

\section{Recurrent Processing}
\label{sec:5}

\subsection{Approach}
\label{sec:6}
Recent work on feedforward convolutional networks has proven the importance of dense sampling and number of the hidden layer units \cite{coates2011analysis}. The question is how a successful feedforward network can be transformed into a computationally not so intense pseudo-recurrent one. In our approach, Coates et al.'s network \cite{coates2011analysis} is adopted and modified to fill-in incomplete (occluded) visual representations (hidden layer activities). The main recurrent processing principles are applied using low complexity operations. The nonlinear dynamical equations that construct the attractors in the high dimensional space are replaced with linear distance comparators. And costly sampling operations such as Markov Chain Monte Carlo (MCMC) \cite{metropolis2004equation} or Gibbs \cite{geman1984stochastic} are replaced with averaging and binary decision operations. In Hopfield networks and Boltzmann machines, learned bidirectional network weights are interpretations of the sensory input, and they are formed during training by iterative energy minimization procedures. In our algorithm, these memories are formed using K-means clustering and linear filtering.

\begin{figure*}
\begin{center}
\fbox{\includegraphics[width=0.9\textwidth]{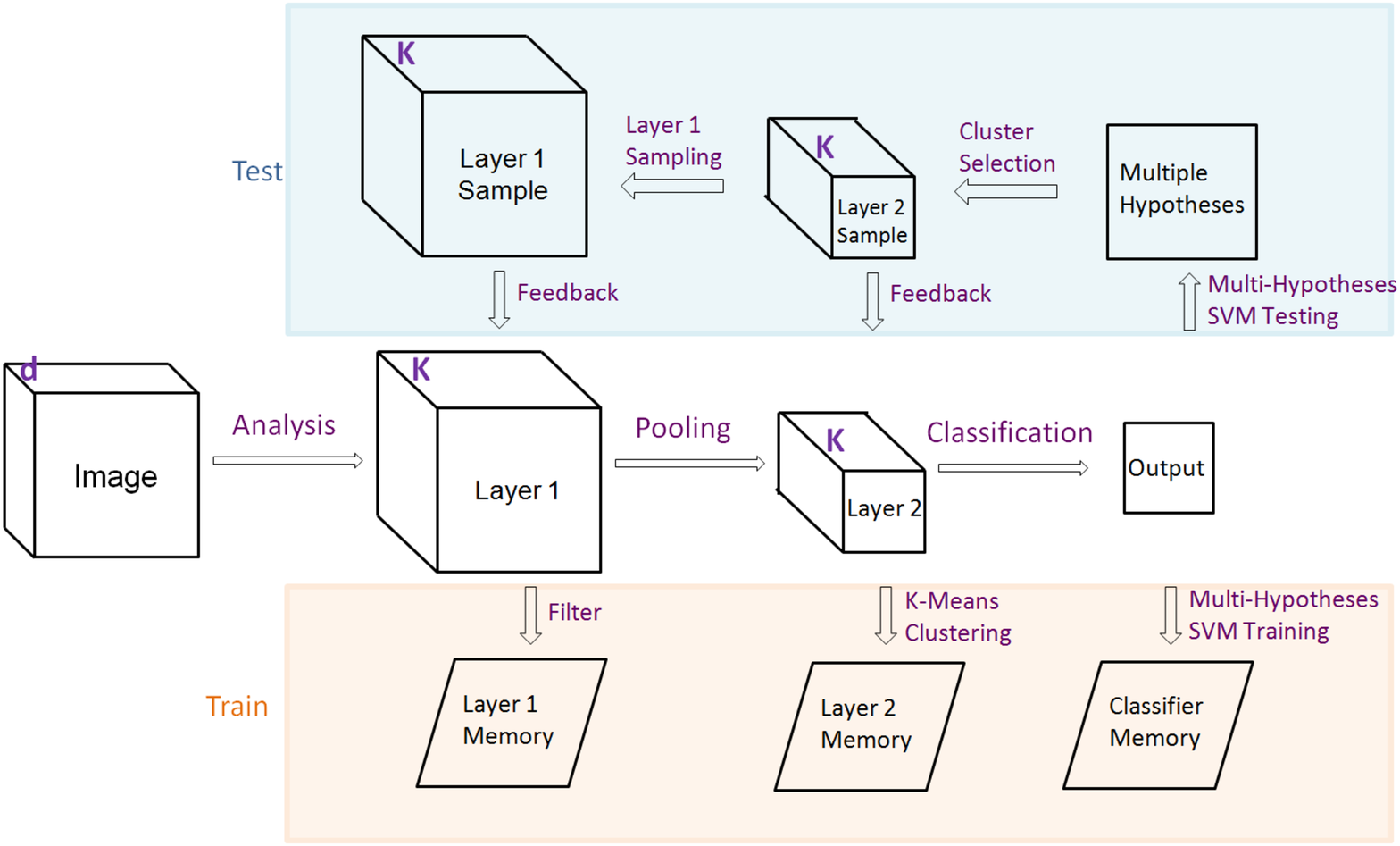}} 
\end{center}
   \caption{The general architecture of the network and the algorithmic stages in training and test stages. The algorithmic stages are shown in arrows and data are shown in boxes. $d$ is the number of channels (i.e. RGB has three) in the image, $K$ is the number of distinct receptive fields in the network. See text for other details on the algorithmic steps.}
\label{fig:fig0}
\end{figure*}


\subsection{Algorithm}
\label{sec:7}
In a recurrent network, the hidden layer activity at time $t$ is given as a function (function $F$, parametrized over  $\theta$ ) of the hidden layer activity ${\mathbf h}$ at time $t-1$ and the current input ${\mathbf x}$ as
\begin{equation}
    {\mathbf h}^{t} = F_{\Theta} ({\mathbf h}^{t-1}, {\mathbf x}^{t}).
\end{equation}
In leaky integration approach, hidden layer activity at time $t-1$ is added for smoother changes,
\begin{equation}
    {\mathbf h}^{t} = \gamma{\mathbf h}^{t-1} + (1-\gamma)F_{\theta} ({\mathbf h}^{t-1}, {\mathbf x}^{t}),
\end{equation}
where $\gamma$ is the leakage from previous hidden layer activity. 

In our framework, we use leaky integration approach and for computational efficiency, costly $F_\theta$ recurrence is replaced with $H^t$ i.e.,
\begin{equation}
    {\mathbf h}^{t} = \gamma{\mathbf h}^{t-1} + (1-\gamma)H^t,
\end{equation}
where $H^t$ is the selected cluster center at time $t$, representing the retrieved network memory for a specific input image. This memory is retrieved such that the distance of previous hidden layer activity $h^{(t-1)}$ to a recorded set of hidden layer activities is minimized, 
\begin{equation}
    H^t = \underset{k} {\mathrm{argmin}} ({\mathbf h}^{t-1} - _y^k\bar{H})^2.
\end{equation}
Here, $\bar{H}$ is the set of the cluster centers, having $K_2$ number of clusters for each class (class decision is $y$). The closest cluster center is utilized ($\underset{k} {\mathrm{argmin}}$ operation) with respect to the class decision $y$ at the output layer \footnote{For simplicity of notation throughout the paper, the argmin operation output is not an index but a selected vector that minimizes the distance between a test vector and a set of target vectors.}. In contrast to KNN imputation approaches \cite{batista2003experimental} in which the search of the best sample is over the whole dataset, vector quantization is utilized in our framework to reduce the computational cost (see also \cite{wang2005support} for utilization of quantization in Support Vector Machines (SVM)). The closest cluster center $H^t$ computation is based on the decision on the class label from the output layer. Therefore, the network uses its class decision to narrow down the set of candidate probability distributions for sampling hidden layer activity. Overall, high level class decision information is used to generate hidden layer activity, that is then merged with the current hidden layer activity. Repeating this procedure in a loop emulates the behavior of a dynamical system, i.e. RNN.   

In the middle row of Figure \ref{fig:fig0}, network architecture is shown as that (1) it has one convolutional hidden layer (which performs Analysis or dimensionality expansion), (2) a subsequent hidden layer computed by Pooling the first hidden layer’s activities in each quadrant and (3) a linear Support Vector Machine (SVM) mimics the output layer of the network and performs multi class Classification on Layer 2 activity (see \cite{coates2011analysis} for more details).

\subsubsection{Training}
\label{sec:8}
During the training phase (lower row of Figure \ref{fig:fig0}), there are 3 stages that are introduced for recurrent processing:

\indent 1. \textbf{Filter}: The first and second hidden layer activities of every training input image is low pass filtered and stored: 
\begin{equation}
    \dot{H}_1 =  \left\{h_1^1, h_1^2,..h_1^N \right\}, \text{ for N training examples and Layer 1},
\end{equation}
\begin{equation}
		\dot{H}_2 =  \left\{h_2^1, h_2^2,..h_2^N \right\}, \text{ for N training examples and Layer 2}.
\end{equation}

\indent 2. \textbf{K-Means Clustering}: Memory formation in an RNN through costly energy minimization is replaced with clustering. The second hidden layer activities ($\dot{H}_2$) are vectorized and clustered using K-means with $K_2$ number of clusters per class (class decision denoted by $y$) or $K_2 \times (Number of Classes)$ number of clusters for non-class specific processing (cf. section \ref{sec:expnumberhypo}). These stored memories are used for sampling followed by imputation during test. Hidden Layer 2 memory set:    
\begin{equation}
    \bar{H}_2^y =  \left\{h_2^1, h_1^2,..h_1^{K_2} \right\},\\
		\bar{\bar{H}}_2 =  \left\{h_2^1, h_2^2,..h_2^N \right\}.
\end{equation}

\indent 3. \textbf{Multi-Hypotheses SVM Training}: In an RNN multiple hypotheses can form and compete with each other to explain sensory data. Multi hypotheses linear classification framework is constructed to imitate this feature. The training is repeated for a subset of data in order to allow multiple hypotheses of the network. This is achieved by excluding the data of a specific single class (eg. Class 1) or a pair of classes (eg. Class 1 and Class 2), and training an SVM for the rest of the data. In the case of single class exclusion, the trained SVM can be used for supplying a second hypothesis. For example, if Class 1 is the first choice of the network that is decided by the “full SVM classifier”, the classifier trained by leaving out Class 1 data is used to give a second hypothesis. In the case of a pair of class exclusions, for example both Class 1 and Class 2 data are left out, the trained SVM gives a third hypothesis, where the first choice is Class 1 and the second choice is Class 2. This collection of classifiers is used during test, to decide which cluster centers of hidden Layer 2 activities will be used for feedback insertion.

S is the SVM classifier for the first choice of the network, $S^p$ is the SVM classifier for the second choice when the first choice was class p and $S^{pq}$ is the SVM classifier for the third choice when the first choice was class p  and second q. 

\subsubsection{Test}
\label{sec:testsubsub}
The test phase (upper row of Figure \ref{fig:fig0}) has the following stages for recurrent processing:\\
\indent 1. \textbf{Pooling}: Test phase starts with the algorithm provided by Coates et al. \cite{coates2011analysis} and computes hidden Layer 2 activity via pooling Layer 1 activity. For test image $i$, at time = $t$:
\begin{equation}
    {\mathbf h}_{2}^{i,t} = P({\mathbf h}_{1}^{i,t}) , \text{ where P is the pooling operation}.
\end{equation}
\indent 2. \textbf{Multi-Hypotheses SVM Testing}: First, second and third class label choices of the network are extracted using the corresponding linear SVM (shown as $S()$ below) in the classifier memory:
\begin{equation}
    {\mathbf y}^{1} = S({\mathbf h}_{2}^{i,t}),\\
		{\mathbf y}^{2} = {S^{y^1}}({\mathbf h}_{2}^{i,t}),\\
		{\mathbf y}^{3} = {S^{y^1 y^2}}({\mathbf h}_{2}^{i,t}).
\end{equation}

\indent 3. \textbf{Sampling}: For each class hypothesis, the cluster centers in the hidden Layer 2 memory ($\bar{H}_2$) which are closest (Euclidian distance) to the current hidden Layer 2 activity are computed. These are hidden layer hypotheses of the network. 3 cluster centers (one for each class hypothesis) are computed as follows:
\begin{equation}
    \widetilde{\mathbf h}_{2,1}^{i,t} = \underset{k} {\mathrm{argmin}} ({\mathbf h}_2^{i,t} - \bar{H}_2^{y^1,k})^2,
\end{equation}
\begin{equation}
		\widetilde{\mathbf h}_{2,2}^{i,t} = \underset{k} {\mathrm{argmin}} ({\mathbf h}_2^{i,t} - \bar{H}_2^{y^2,k})^2,
\end{equation}
\begin{equation}
		\widetilde{\mathbf h}_{2,3}^{i,t} = \underset{k} {\mathrm{argmin}} ({\mathbf h}_2^{i,t} - \bar{H}_2^{y^3,k})^2.
\end{equation}

\indent 4. \textbf{Competition}: In a “winner-takes-all” configuration, the closest of the clusters computed above is chosen as the Layer 2 hidden activity sample:
\begin{equation}
    \widetilde{\mathbf h}_{2,A}^{i,t} = \underset{m} {\mathrm{argmin}} (\widetilde{\mathbf h}_{2,m}^{i,t} - {\mathbf h}_{2}^{i,t})^2.
\end{equation}
For the “average” configuration, the average of the $m$ clusters (for each class) is assigned as the sample:
\begin{equation}
    \widetilde{\mathbf h}_{2,A}^{i,t} = \frac{1}{m}\displaystyle\sum_{n=1}^{m}(\widetilde{\mathbf h}_{2,m}^{i,t}).
\end{equation}
For non-class specific configuration, instead of computing 3 closest centers for each of the class hypotheses, 3 closest clusters are computed regardless of class hypotheses. Another hidden Layer 2 memory set ($\bar{\bar{H}}_2$, see above) is used for distance computation:
\begin{equation}
    \widetilde{\mathbf h}_{2,A}^{i,t} = \underset{k} {\mathrm{argmin}} ({\mathbf h}_2^{i,t} - \bar{\bar{H}}_2^{k})^2,
\end{equation}

\indent 5. \textbf{Feedback (Layer 2)}:The Layer 2 sample is merged (feedback magnitude, $\alpha$) with the test image Layer 2 activity, to generate hidden layer activity at time $t+1$:
\begin{equation}
    {\mathbf h}_{2}^{i,t+1} =  \frac{1}{1 + \alpha} ({\mathbf h}_2^{i,t} + \alpha \widetilde{\mathbf h}_{2,A}^{i,t}).
\end{equation}

\indent 6. \textbf{Layer 1 Sampling}: The modified hidden Layer 2 activity is used to compute the most similar training set image, using the Euclidean distance:
\begin{equation}
    {\mathbf L}_{2}^{i,t} =  \underset{k} {\mathrm{argmin}} ({\mathbf h}_2^{i,t+1} - \dot{H}_2^{k})^2,\\
		\text{which is the index of the most similar training data}.
\end{equation}
The hidden Layer 1 activity of the most similar training data is fetched from the Layer 1 memory, as the Layer 1 sample of the network:
\begin{equation}
    \widetilde{\mathbf h}_{1,A}^{i,t} = \dot{H}_1^{L},
\end{equation}
 \indent 7. \textbf{Feedback (Layer1)}: The Layer 1 sample is merged (feedback magnitude $\beta$) with the test image Layer 1 activity, to generate hidden layer activity at time $t+1$. 
\begin{equation}
    {\mathbf h}_{1}^{i,t+1} =  \frac{1}{1 + \beta} ({\mathbf h}_1^{i,t} + \beta \widetilde{\mathbf h}_{1,A}^{i,t}).
\end{equation}

\indent 8. \textbf{Pooling (again)}: Layer 1 activity is pooled to compute Layer 2 activity. Then, this activity is averaged (feedback ratio, $\tau$) with previously computed Layer 2 activity.
\begin{equation}
    {\mathbf h}_{2}^{i,t+1} :=  \frac{1}{1 + \tau} ({\mathbf h}_2^{i,t+1} + \tau P({\mathbf h}_{1}^{i,t+1})).
\end{equation}

This procedure is repeated for multiple iterations starting from the second stage. The feedback magnitude is halved at each iteration for simulated annealing purposes. At the end of iterations, ${\mathbf y}^{1}$ is given as the classification output of the algorithm.

\section{Experiments}
\label{sec:9}

\begin{figure*}
\begin{center}
\fbox{ \includegraphics[width=0.9\textwidth]{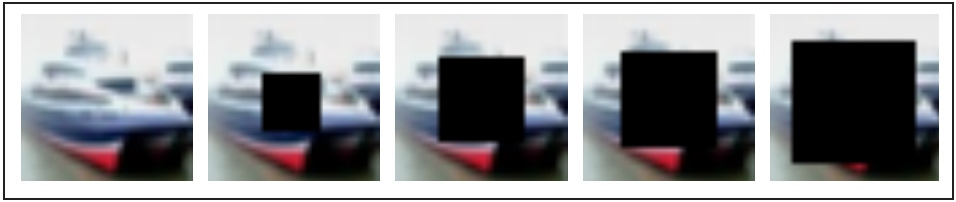}}
\end{center}
   \caption{Sample images used in the experiments. Original image, 11, 25, 33 and 50 percent occlusions are demonstrated.}
\label{fig:fig0p5}
\end{figure*}

\begin{figure*}
\begin{center}
\fbox{ \includegraphics[width=0.9\textwidth]{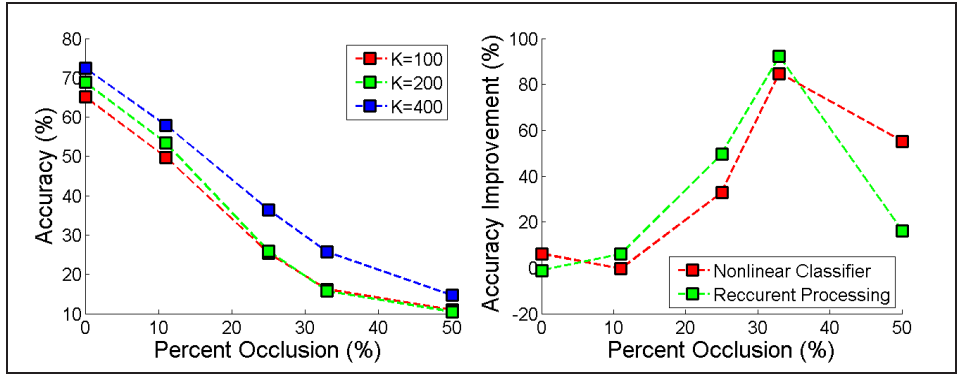}}
\end{center}
   \caption{\textbf{Left}: The performance of the feedforward network in classifying occluded objects in CIFAR 10 dataset, for different numbers of hidden layer units shown separately.  \textbf{Right}: The accuracy improvement when linear SVM is replaced with RBF kernel nonlinear SVM and when pseudo-recurrent processing (only Layer 2 feedback) is applied.}
\label{fig:fig1}
\end{figure*}

For performance evaluation, the CIFAR 10 dataset’s \cite{krizhevsky2009learning} test batch is modified to simulate occlusions (Figure \ref{fig:fig0p5}). The middle of the images is deleted, i.e., filled in with zeros, for various area-wise occlusion percentages. Occlusion pattern is assumed unknown. The reason for occluding the middle part is for making the case harder: the effect of occlusion is more or less distributed to hidden layer activities due to pooling. Therefore we do not present results on other easier occlusion locations (eg. upper left corner). The occlusion manifests more as the distortion in the hidden unit activities as opposed to missing data. However, due to the uniform nature of the occluder, the distortion is still localized to a subset of the receptive fields. Fifty thousand training images and ten thousand test images are used for the experiments. The performance of the original algorithm \cite{krizhevsky2009learning} on the occluded test images are shown in Figure \ref{fig:fig1}  (left), for three different numbers of hidden layer units. The accuracy drops down to chance level for 50\% occlusion. 

\begin{figure*}
\begin{center}
\fbox{ \includegraphics[width=0.98\textwidth]{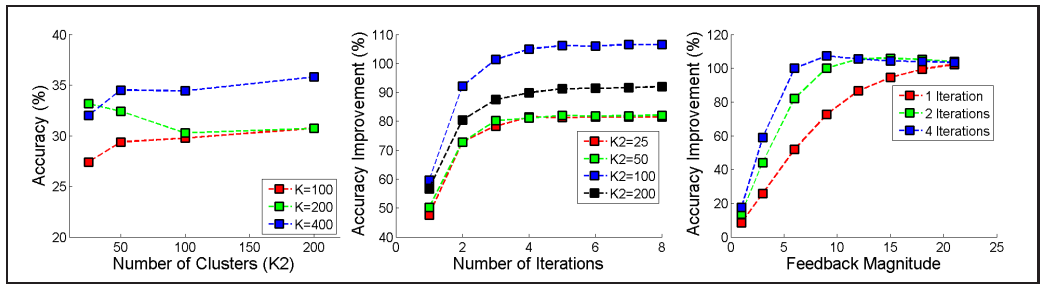}}
\end{center}
   \caption{ \textbf{Left}: The accuracy of the algorithm in 33\% occlusion case as a function of number of clusters per class ($K_2$). Middle: Percent accuracy improvement due to recurrent processing as a function of number of feedforward-feedback iterations, for different $K_2$ values. \textbf{Right}: Percent accuracy improvement due to recurrent processing as a function of feedback magnitude, for different number of iterations.}
\label{fig:fig2}
\end{figure*}

\begin{figure*}
\begin{center}
\fbox{ \includegraphics[width=0.9\textwidth]{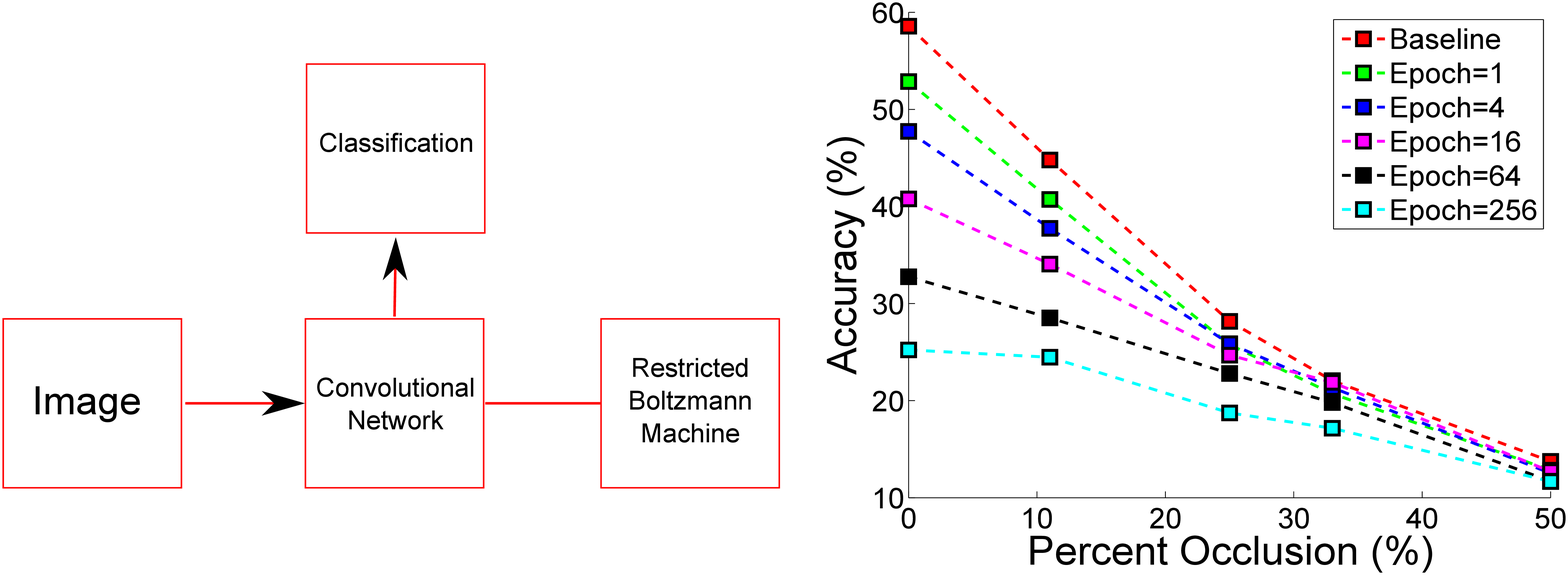}}
\end{center}
   \caption{\textbf{Left}: In these experiments, the hidden layer activities of the convolutional network is used as input to the Restricted Boltmann Machine (RBM). The Gibbs sampling on RBM visible units is used instead of the feedback processing algorithm proposed in this paper. After some number of Gibbs epochs, the regularized convolutional hidden layer activity is used for classification.  \textbf{Right}: The effect of Gibbs sampling on RBM visible units for different number of Gibbs epochs. The RBM processing does not improve occluded object classification, moreover it is detrimental for un-occluded objects.}
\label{fig:fig5}
\end{figure*} 

In the first set of experiments (\ref{sec:11}, \ref{sec:rbm}, \ref{sec:12}, \ref{sec:expnumberhypo}), only Layer 2 sampling and feedback is executed (steps 6-7-8-9 are skipped) to test the sole effect of higher level feedback into the network.    

\subsection{Nonlinear vs. Recurrent Processing}
\label{sec:11}
The accuracy improvement due to recurrent processing is shown in Figure \ref{fig:fig1} (right) (no occlusion in training data, but occlusion in the test data). We emphasize that the recurrent processing does not impair performance in zero occlusion case. It improves performance as much as 2x for 33\% occlusion case. When compared with the performance of a nonlinear SVM \cite{chang2011libsvm} (RBF kernel, hidden layer activities as the feature space, grid search for best SVM parameter), it is observed that the recurrent processing exceeds the performance of nonlinear SVM (for most occlusion cases) but with only a fraction of the computational cost at test time: 13 sec nonlinear SVM (C++ implementation) vs 1 sec recurrent processing (Matlab implementation). A comparable implementation of the algorithm and systematic timing experiments are planned as future work but still we can make complexity comparisons. The computational complexity of the nonlinear SVM is $\mathcal{O}(ND)$  where $N$ is the number of hidden layer neurons, and $D$ is the number of support vectors. The complexity of the proposed algorithm is $\mathcal{O}(NK_{2})$ where $K_2$ is the number of cluster centers learned during training. The experiments indicate huge savings: $D$ is orders of magnitude larger than $K_2$. 

It is observed that nonlinear SVM does not outperform linear SVM for zero occlusion case but improves performance for occluded objects. Hence, performance-wise, recurrent processing upgrades linear SVM into nonlinear SVM, using small amount of computational resources. The theoretical and experimental investigation of nonlinear SVM's power in incomplete data classification is out of the scope of this paper and should definitely be pursued.      

\subsection{Restricted Boltzmann Machines Under Unknown Occlusion Mask}
\label{sec:rbm}
Both in inpainting studies \cite{bertalmio2000image,roth2005fields} and Restricted Boltzmann Machine (RBM) experiments \cite{salakhutdinov2012efficient}, it is assumed that the region of occlusion is known and a mask is given to the algorithm, after which the missing information is generated through a sampling process. We experimented with the data generation capability of RBMs when the location of the occlusion is unknown. In order to make a valid comparison, we used binarized hidden layer activities of the convolutional network as the input to a classical RBM network (Figure \ref{fig:fig5}, left), instead of the image. In this configuration RBM learns the statistics of the convolutional network activities. We trained a single layer fully connected RBM with 800 hidden neurons (K=200, batch size 100, learning rate 0.1, with 100 training epochs). RBM network as a generative engine, is intended to replace the recurrent processing proposed in this study by applying alternating Gibbs sampling process \cite{geman1984stochastic}. In this procedure the visible layer of the RBM (that receives hidden layer activity of convolutional network) is initialized with the occluded test input and Gibbs sampling is expected to correct/regularize the distorted data. We experimented with different numbers of Gibbs epochs (Figure \ref{fig:fig5}, right), and observed that RBM is not able to improve performance of classification for occluded objects. Moreover, serious performance drop was observed for non-occluded (original) test set. The failure of RBM can be explained by unsubtle changes in the visible units during sampling and large unavoidable reconstruction errors (around 70, that is almost 10\% of the signal). Consequently, the erratic changes in the visible units during sampling impairs more than what can be gained by the corrections on the corrupted data. This distortion is more dramatic when there is no occlusion in the image. We conclude that RBM is not suitable for occlusion handling when the the location of occlusion is not provided.       

\subsection{Number of Clusters, Iterations and Magnitude}
\label{sec:12}
For analyzing our network to various choices of the parameters, we conducted several experiments (for 33\% occlusion case). Number of clusters ($K_2$) is varied and the results (Figure \ref{fig:fig2}, left) show that the performance is saturated at around 50 cluster centers per class. Number of feedforward-feedback iterations is examined and the results (Figure \ref{fig:fig2}, middle) indicate that 3 iterations are enough to reach maximum accuracy improvement. The feedback magnitude ($\alpha$) is cross-varied with number of iterations in another set of batch runs (Figure \ref{fig:fig2}, right). It is observed that a large feedback magnitude with only one iteration can attain the same performance gain of larger number of iterations with lower feedback magnitude, and iteration number is an important parameter for real-time operation concerns. However, a large feedback magnitude is observed to be more detrimental for zero occlusion case.     

\begin{figure*}
\begin{center}
\fbox{ \includegraphics[width=0.9\textwidth]{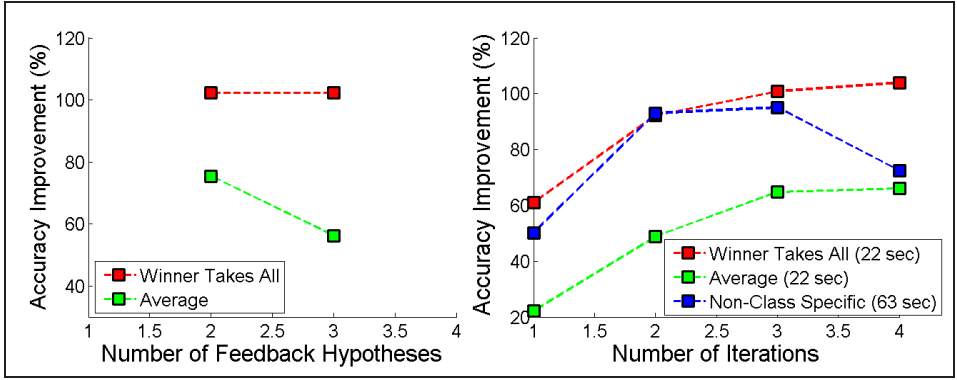}}
\end{center}
   \caption{\textbf{Left}: Accuracy improvement due to recurrent processing for two different sampling schemes, winner-takes-all and averaging. \textbf{Right}: Accuracy improvement due to recurrent processing in non-class specific sampling and class specific feedback (winner-takes-all and averaging) are shown as a function of number of iterations.}
\label{fig:fig3}
\end{figure*} 

\begin{figure*}
\begin{center}
\fbox{ \includegraphics[width=0.9\textwidth]{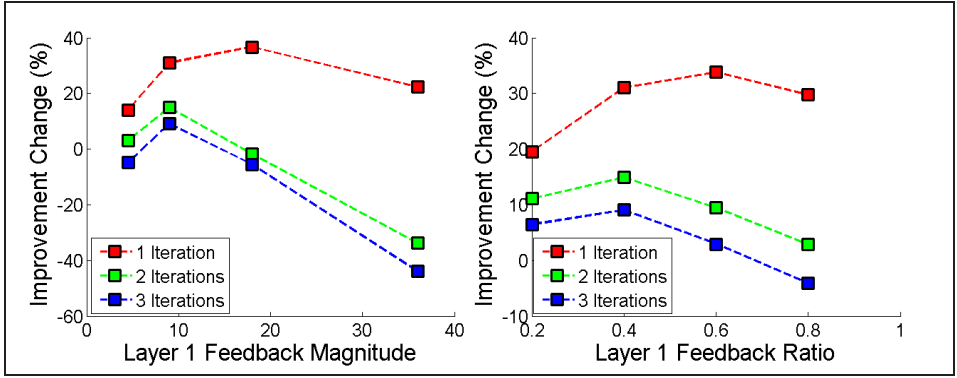}}
\end{center}
   \caption{\textbf{Left}: Accuracy improvement change (wrt to Layer 2 feedback improvement) due to Layer 1 feedback as a function of Layer 1 feedback magnitude. \textbf{Right}: Accuracy improvement change due to Layer 1 feedback as a function of Layer 1 feedback ratio.}
\label{fig:fig4}
\end{figure*} 

\subsection{Number of Hypotheses and Competition}
\label{sec:expnumberhypo}
Accuracy improvement due to recurrent processing is tested for two different sampling schemes (for 33\% occlusion case): winner-takes-all and average. In winner-takes-all, the cluster center closest to the current image hidden layer activity is chosen as the Layer 2 sample among the cluster centers from computed class hypotheses. Accordingly, only one cluster wins to be fed-back into the hidden layer activity (see \cite{zheng2013winner} for the effect of winner-takes-all behavior in neural networks). In average scheme, the cluster centers from different hypotheses are averaged and fed back into the hidden Layer 2 activity. The accuracy improvement results (Figure \ref{fig:fig3}, left) show inferiority of averaging scheme, getting worse with the number of class hypotheses. Another sampling scheme is using non-class specific cluster centers that are learned from the whole data. In this scheme, the hidden layer activity is assumed to be disconnected from output layer decisions, and feedback is generically computed by finding the closest cluster center in non-class specific cluster memory. The accuracy improvement results (Figure \ref{fig:fig3}, right) show the superiority of class specific (winner-takes-all) feedback, both in terms of performance and speed. Specifically, non-class specific feedback impairs classification performance as the feedforward-feedback iterations proceeds. 

\subsection{Recurrent Processing in Hidden Layer 1}
\label{sec:13}
In order to investigate the effect of feedback onto lower layers of the network, Layer 1 sampling procedures (stage 6 explained in section \ref{sec:testsubsub}) are executed and the Layer 1 sample is fed back into the network hidden layer activity (stage 7). It should be noted that Layer 1 feedback is a more costly operation (for both memory and processing) than Layer 2 feedback, because of a large memory of Layer 1 hidden layer activities and the need for pairwise distance computations to all training data (stage 6). 

The effect of Layer 1 feedback magnitude β and feedback ratio τ is examined in a set of experiments. The results (Figure \ref{fig:fig4}) show that Layer 1 feedback gives marginal accuracy gain compared to Layer 2 feedback. The gain is present especially for small feedback magnitude condition in Layer 2 and only in the first iteration, for which there is still room for improvement. However, the net effect of Layer 1 feedback is close to zero (or negative) for larger number of iterations. Layer 1 feedback seems to distort the hidden layer activities too much, impairing classification performance after first feedback iteration. However, one iteration of combined Layer 1 and Layer 2 feedback gives the same performance of multiple only Layer 2 feedback iterations.

\begin{figure}
\begin{center}
\fbox{ \includegraphics[width=0.9\textwidth]{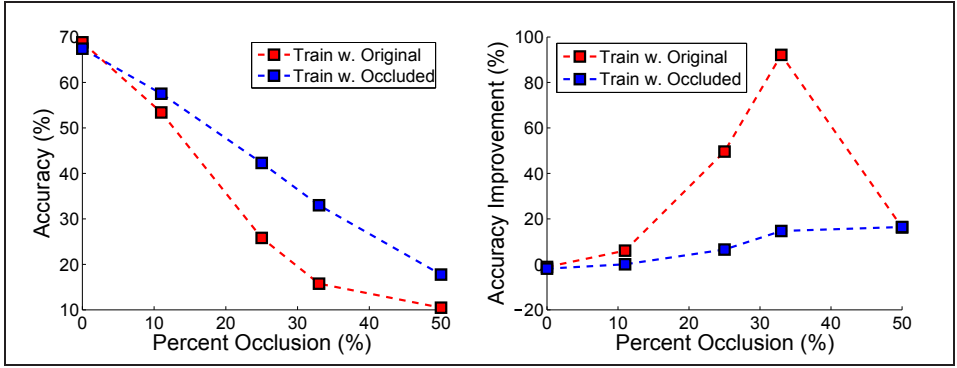}}
\end{center}
   \caption{The effect of using occluded images in the training data. We included occluded versions of every training image (11, 25, 33 and 50 percent occlusions) to the training image set.  \textbf{Left}: The classification accuracy with respect to occlusion levels, for original training set and augmented set. \textbf{Right}: The percent improvement in classification accuracy with respect to occlusion levels for for original training set and augmented set.}
\label{fig:fig6}
\end{figure}

\subsection{Training with Occluded Images}

Artificially expanding the data set using image transformations have been successfully used for reducing overfitting \cite{sanchez2011high,ciresan2012multi,krizhevsky2012imagenet}. We tested this approach for improving classification performance on occluded objects. Training image set is expanded by including occluded versions of the images, all 4 levels of occlusion (11, 25, 33, 50). To the best of our knowledge, this type of augmentation was not reported in the literature before. The receptive fields and the cluster centers are unchanged, but only the SVM classifiers are trained on the new dataset. The results show a great improvement in classification accuracy due to this data augmentation (Figure \ref{fig:fig6}, left). Within this configuration, feedback algorithm still enhances performance but its impact is reduced (Figure \ref{fig:fig6}, right). 

\begin{figure}
\begin{center}
\fbox{ \includegraphics[width=0.9\textwidth]{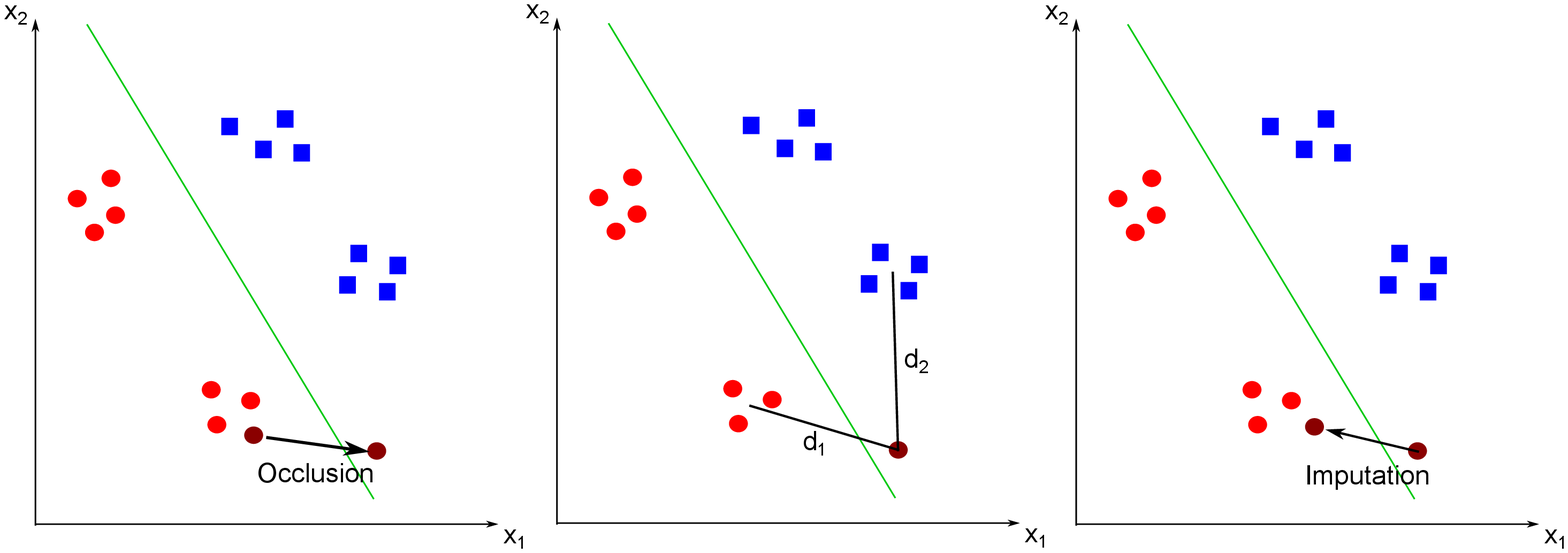}}
\end{center}
   \caption{The toy example to illustrate the mechanism of imputation. There are two classes (rectangles and circles) and two feature dimensions. The classes are composed of distinct clusters. \textbf{Left}: Visual occlusion distorts a subset of the feature attributes (dark red point), and sweeps the data onto another class region, which causes incorrect classification. \textbf{Middle}: However, for some of these distortions the closest cluster to the new data point still belongs to the correct class. \textbf{Right}: If the data point is moved towards the closest cluster center, it may land onto the correct class region and saved from incorrect classification.}
\label{fig:fig7}
\end{figure} 

\section{Analysis}
\label{sec:14}
Correction of corrupted hidden layer activity with closest cluster center imputation improves classification performance significantly, especially for medium level occlusions. The mechanisms of this improvement is illustrated in Figure \ref{fig:fig7}. In our toy example, feature dimension and number of classes are both two, and we assume that each class is composed of distinct clusters. This assumption is valid for ecological object classification. Visual occlusion distorts a subset of the feature attributes (dark red point), and sweeps the data onto another class region (Figure \ref{fig:fig7}, left) which causes incorrect classification. However, for some of these distortions, the closest cluster to the new data point still belongs to the correct class (Figure \ref{fig:fig7}, middle). If the data point is moved towards the closest cluster center, it may land onto the correct class region and saved from incorrect classification (Figure \ref{fig:fig7}, right). By moving the data towards the cluster mean, the log likelihood (Gaussian assumption) of the correct class (c1) is improved:
\begin{equation}
    {\frac {p_{c1}(x)}{p_{c2}(x)}} = \frac {(x - \mu_{c1})^2}{(x - \mu_{c2})^2}, \text{   where x is the data vector}, \mu_{c1} \text{ and } \mu_{c2}  \text{ are cluster centers}. \\
\end{equation}

In this approach, the cluster nature of the data and the locality of adversary effect of the occlusion is exploited, however the illustrated correction is not guaranteed. The probability of correction drops for severe occlusions (the data moves very far away from the correct cluster) and for more uniform distribution of the data (closest cluster center is more likely to belong to the incorrect class). Localization of occlusion is beneficial for more accurately estimating the closest cluster center, such that once the occlusion is localized to a subset of the feature attributes the 'healthy' attributes give a more reliable distance measure to the cluster centers. Therefore it is essential that an occlusion localization is devised for further improvement in occluded object classification.    

\section{Computational Complexity}
\label{sec:ComputComplex}
A comparable implementation of the algorithm and systematic timing experiments are planned as future work but still we can make complexity comparisons. The complexity of the proposed algorithm is $\mathcal{O}(NK_{2})$ where $N$ is the number of hidden layer neurons and $K_2$ is the number of cluster centers learned during training. The computational complexity of the nonlinear SVM is $\mathcal{O}(ND)$  where $D$ is the number of support vectors. Whereas the computational complexity of Restricted Boltzmann Machines are $\mathcal{O}(N^2 K^2)$, in which $K$ is the number of hidden layer components. First observation is that $D$ is orders of magnitude larger than $K_2$ for interesting problems in computer vision. Secondly we can infer that RBMs are computationally much demanding than our approach by looking at the big-oh complexity. There are GPU implementations of RNN algorithms that can be trained and tested on large datasets, but compensation via hardware and software speed-ups does not alleviate the inherent complexity of these algorithms.

\section{Discussion and Future Work}
\label{sec:15}
In our work we attempted to alleviate the impairment caused by visual occlusions, and we adopted a blind approach: the location of the occlusion is unknown. To our knowledge, this is the first recurrent neural network study that systematically examines the effect of occlusion on classification performance in a standard image dataset and in this blind setting. In recurrent neural network literature, it is assumed that the occlusion is detected and localized as a preprocessing step, but in general it is computationally expensive and a gold standard occlusion localization approach is nonexistent. We introduced a recurrent algorithm that imputes the hidden layer activities of a convolutional neural network, without resorting to the occlusion detection step. Iteratively merging hidden layer activities with samples from a mixture model improved classification accuracy for occluded objects up to 100\%. Selecting a single sample is shown to work better than averaging many class hypotheses or non-class specific selection. Feedback to lower layers has shown diminishing return. 

In our experiments, there is no variability on the location of the occluder: it is always in the center of the image. This is chosen to make the case as hard as possible for the algorithm. Yet, it should be noted that the algorithm does not use any information or an ad-hoc procedure to exploit this invariance in location. Thus it is possible to assert that the algorithm works in the worst possible scenario of occlusion location.

We tested the performance of RBMs in this blind setting, and observed that Gibbs sampling in the RBM visible units is not capable of restoring distorted feature attributes. Hence, energy based recurrent networks seem to require a localization stage in order to accurately classify occluded objects.

Moreover we experimented executing neural network training with occluded images, which is ecologically more natural. It is observed that the performance is enhanced significantly when some of the training images also suffer from occlusions. To the best of our knowledge, this type of data augmentation was not reported in the literature and this observation is novel. 

As a future work, a second convolutional layer can be introduced after Layer 1 of the proposed network using unsupervised methods in determining the local receptive fields \cite{coates2011selecting}. In that case it is possible to devise more plausible inter-layer feedback from second convolutional layer to the first. This will extend the scope of pseudo-recurrent processing and will allow more complicated multi-layer feedforward-feedback loops. Also, detection and localization of occlusion seems beneficial, especially for hidden Layer 1 feedback, and devising a fast and effective occlusion detection needs further development. Moreover, the nonlinear SVM's power for incomplete data classification and data augmentation with occluded images seem very promising and need further investigation.

\section{Acknowledgments}
This research is supported by The Scientific and Technological Research Council of Turkey (TUB\.{I}TAK) Career Grant, No: 114E554 and Turgut Ozal University BAP Grant, No: 006-10-2013.

\newpage
\section*{References}

\bibliography{OzgurYilmaz_NeuralComputingAndApplications_2014_Bib}   

\end{document}